\documentclass[journal,10pt]{IEEEtran}
\ifCLASSINFOpdf
\else
\fi
\usepackage{comment}

\usepackage{epsfig}
\usepackage{graphicx}
\usepackage{amsmath}
\usepackage{amssymb}

\usepackage[linesnumbered,ruled,vlined]{algorithm2e}

\usepackage{amsmath,graphicx}

\usepackage{pstricks,graphicx}
\usepackage{setspace}
\usepackage{psfrag}

\usepackage{multirow}

\usepackage{amsmath,graphicx}

\usepackage{pstricks,graphicx}
\usepackage{setspace}
\usepackage{psfrag}
\usepackage{amssymb}
\usepackage{amsmath}
\usepackage{amsthm}

\newtheorem{remark}{Remark}

\begin{document}



\title{Approximated Orthonormal Normalisation in Training Neural Networks}

\graphicspath{{figures/}}
%

\author{Guoqiang~Zhang,  Kenta Niwa and W. B. Kleijn
\thanks{G.~Zhang is with the School of Electrical and Data Engineering, University of Technology, Sydney, Australia. Email: {guoqiang.zhang@uts.edu.au}}
\thanks{K.~Niwa is with Nippon Telegraph and Telephone (NTT) Corporation, Japan.
Email: {niwa.kenta@lab.ntt.co.jp}}
\thanks{W. B. Kleijn is with the School of Engineering and Computer Science, Victoria University of Wellington, New Zealand. Email: {bastiaan.kleijn@ecs.vuw.ac.nz}}
}

\maketitle

\begin{abstract}
Generalisation of a deep neural network (DNN) is one major concern when employing the deep learning approach for solving practical problems. In this paper we propose a new technique, named approximated orthonormal normalisation (AON), to improve the generalisation capacity of a DNN model. Considering a weight matrix $W$ from a particular neural layer in the model, our objective is to design a function $h(W)$ such that its row vectors are approximately orthogonal to each other while allowing the DNN model to fit the training data sufficiently accurate. By doing so, it would avoid co-adaptation among neurons of the same layer to be able to improve network-generalisation capacity.  Specifically, at each iteration, we first approximate $(WW^T)^{-\frac{1}{2}}$ using its Taylor expansion before multiplying the matrix $W$.  After that, the matrix product is then normalised by applying the spectral normalisation (SN) technique to obtain $h(W)$.  Conceptually speaking, AON is designed to turn orthonormal regularisation into orthonormal normalisation to avoid manual balancing the original and penalty functions. Experimental results show that AON yields promising validation performance compared to orthonormal regularisation. 
\end{abstract}



\section{Introduction}

How to train a deep neural network  (DNN) to maximize its generalisation capacity has been a challenging task.  The training process may be affected by various factors such as the nature of nonlinear activation functions,  weight initialisation,  neural network architectures, and optimization methods like stochastic gradient descent (SGD). In the past few years, different techniques have been proposed to improve the training process from different perspectives. Considering selection of the activation function,  the rectified linear unit (ReLU) was found to be much more effective than the binary unit in feed-forward neural networks (FNNs) and convolutional neural networks (CNNs)  \cite{Nair10ReLU}. Careful weight initialization based on the properties of the activation function and layerwise neuron-number has also been found to be essential for effective training (e.g., \cite{He15WeiInti, Sutskever13NAG}). Nowadays, neural networks with shortcuts (e.g., ResNet \cite{He15ResNet, Zagoruyko16WideResNet}, DenseNet \cite{Huang17DenseNet}, and Unet \cite{Ronneberger15Unet}) become increasingly popular as introduction of the shortcuts greatly alleviates the issue of gradient vanishing or explosion, which become severe issues when training extremely deep neural networks. From the optimization point of view, SGD with momentum is empirically found to produce DNNs with good generalisation capacity over other gradient based methods (e.g., Adam \cite{Kingma17, Reddi18Amsgrad}, AdaGrad \cite{Duchi11AdaGrad},  RMSProp \cite{Tieleman12RMSProp}).

In recent years, a family of normalisation techniques have been proposed to accelerate the training process and produce high quality DNN models. The motivation behind these techniques is to make proper adjustment at each individual layer so that either the input or output statistics of the activation functions of the layer are unified in terms of the first and/or second moments. By doing so, the problem of internal covariance shift can be largely alleviated, thus significantly improving the efficiency of the back-propagation optimisation methods. Those techniques can be roughly classified as (a): data-driven normalisation, (b): activation-function normalisation, and (c): weight-driven normalisation. 

We now briefly review the above three normalisation techniques. 
Data-driven normalisation operates directly on the  layer-wise internal features of training data, which includes for example batch normalisation \cite{Ioffe15BN, Ioffe17BRN}, layer normalisation \cite{Ba16LN}, instance normalisation \cite{Ulyanov17InstanceNorm}, group normalisation \cite{Wu18GroupNorm}, decorrelated batch normalisation (DBN) \cite{Huang18DBN}, and iterative normalisation (IterNorm) \cite{Huang19IterNorm}. This type of normalisations was shown to be remarkably effective but one often has to carefully handle the inconsistency between training and inference, as the input statistics at the inference stage might be changed due to a reduced number of input samples.
 Activation-function normalisation intends to design proper activation functions that are able to keep certain statistics unchanged between its input and output  \cite{Klambauer17SNN}. Weight-driven normalisation indirectly regulate the statistics of the layer-wise internal features by building and implictly imposing constraints on the weight matrices of the neural layers, which include weight normalisation (WN) \cite{Salimans16WN},  centered-weight normalisation (CWN) \cite{Huang17CWN}, and spectral normalisation (SN) \cite{Miyato18SN}. It is reported in \cite{Huang17CWN, Hoffer18N} that CWN (or WN) combined with BN often provides better performance than BN alone. SN is shown to be effective when training generative adversarial networks (GANs) \cite{Miyato18SN}.  
 
Besides weight-driven normalisation, different weight regularisation techniques have also been proposed in the literature. The basic idea is to add specific penalty functions of the weight matrices to the original objective function when training the DNN model to influence the behaviours of the weight matrices. The weight decay is one popular technique, which poses a quadratic weight penalty function. In \cite{Brock16orth},  orthonormal regularisation is proposed for pushing the vectors in each weight matrix to be mutually orthogonal with their norms being pushed close to one.  We will briefly review orthonormal regularisation in Subsection \ref{subsec:pre} later on to motivate our new normalisation technique. Variations of orthonormal regularisation can be found in recent works \cite{Jia19ODNN, Bansal18OR}.
 


In this work, we develop a new weight normalisation method, termed as approximated orthonormal normalisation (AON), to improve the generalisation capacity of DNNs.  Suppose $W$ is a weight matrix extracted from a neural layer. AON attempts to construct a function $h(W)$ such that its row vectors are approximately orthogonal. By doing so, it prevents the constructed weight matrix $h(W)$ from having low rank and from overlearning the training data, which leads to better generalisation of the resulting DNN model. To start with, an approximation of $(WW^T)^{-\frac{1}{2}}$ is obtained using its Taylor expansion before multiplying $W$.  The matrix product is then normalised by SN to obtain $h(W)$. The SN technique is employed to ensure that $h(W)$ would never grow out of control. Differently from orthonormal regularisation, AON smoothly embeds the matrix-orthonormal property to the considered DNN model, which makes the training procedure both cleaner and simpler than using the regularisation technique.   Experimental results on training VGG16 \cite{Simonyan16DCNN} for CIFAR10 and CIFAR100 show that AON consistently outperform orthonormal regularisation with noticeable gains.

\section{Related Work}

Our work is partially motivated by the development of SN in the literature. The authors of SN firstly proposed a spectral-norm regularisation in their earlier work \cite{Yoshida17SNR}, which, as the name suggests,  adds a penalty term to the objective function to implictly regularise the spectral norm of a weight matrix from a DNN model. In general, it is rather difficult  for the regularisation approach to set the spectral norm to a designated value without manual parameter-tuning. Later on, they proposed SN in \cite{Miyato18SN}, which allows to explicitly set the spectral norm of a weight matrix to a designated value without introducing any additional penalty term.  Their basic idea is to first compute an approximation of the spectral norm per iteration and then normalise the weight matrix by dividing it with the obtained approximated spectral norm. 

We note that orthonormal regularisation (see \cite{Brock16orth}) also attempts to add a penalty term to the objective function to make the row vectors of a weight matrix roughly orthonormal. We follow a similar design principle as SN to turn orthonormal regularisation into orthonormal normalisation. By doing so, no penalty term needs to be introduced, making the training procedure considerably simpler.  

Our work is also related to the development of DBN \cite{Huang18DBN} and IterNorm \cite{Huang19IterNorm} as extensions of BN. It is noted that BN centers and scales the input to each neuron within a mini-batch but does not consider the correlations among the inputs of the neurons in the same layer. DBN extends BN by whitening the covariance matrix of the internal features of a neural layer by performing eigen-decomposition or singular value decomposition (SVD).  In general,  DBN improves over BN w.r.t. both training efficiency and generalisation capacity, but it relies heavily on a large batch size and requires high computation complexity. Later on, the authors of DBN found in  \cite{Huang19IterNorm} that full-whitening does not always improve generalisation capacity, especially for small batch size. Based on their observations, they proposed IterNorm to perform approximate-whitening instead, which was empirically found to outperform DBN.   

While IterNorm conducts approximate-whitening of the covariance matrices over internal features of input data, AON intends to make the constructed matrix $h(W)$ have approximately orthonormal row vectors. That is, IterNorm and AON operate on data and weight matrices, respectively. As will be shown later on,  another difference is that IterNorm makes use of Newton's iteration to approximate the whitening matrix, while AON utilises Taylor expansion to obtain approximate orthonormality of the row vectors of $h(W)$. 

\section{Approximated Orthonormal  Normalisation}
\subsection{Preliminary}
\label{subsec:pre}

{\sloppypar Suppose we have a sequence of $L$ pairs of training samples $\{(x_i, y_i)| i=1,\ldots, L\}$, where $x_i$ and $y_i$ represent the input and output, respectively. {To start with, we consider training a fully connected neural network with the weights $\{W_i|i=1,\ldots,N\}$ of $N$ layers.}\footnote{One can extend the work to include the bias parameters.}{}  With the considered DNN model, each sample $x_i$ undergoes a sequence of matrix multiplications and nonlinear functional operations to yield prediction of $y_i$.  The objective is to find the proper weights $\{W_i\}$ so that the  network maps the input $\{x_i\}$ to the output $\{y_i\}$ accurately. Mathematically, the training procedure intends to solve a highly nonlinear and nonconvex optimization problem of the form
\begin{align}
\min_{\{W_i\}} \sum_{i=1}^{L} \mathrm{dis}(f_N(\ldots f_2(W_2f_1(W_1x_i))), y_i) \hspace{-0.5mm}+\hspace{-0.5mm} \beta \sum_{i}^N p(W_i),  \label{equ:optimization}
\end{align}
where $\mathrm{dis}(\cdot, \cdot)$ denotes the distance measure between the network prediction for the sample $x_i$ and its ground truth $y_i$, $f_i$ denotes nonlinear activation function at layer $i$, and $\beta$ is a scalar coefficient. The 2nd term in (\ref{equ:optimization})  represents a regularization penalty function of the weight matrices. For the well-known weight decay technique \cite{Krogh92weightdecay}, $ p(W_i)$ becomes a quadratic penalty function of $W_i$, which prevents the weight matrices from growing out of control.  }
 
Next we briefly review the orthonormal regularisation proposed in \cite{Brock16orth}, of which the penalty function for a weight matrix $W$ of a neural layer takes the form of
\begin{align}
p_{orth}(W) = \frac{1}{m^2} \|WW^T - I \|_2^2, \label{equ:orthnorm}
\end{align}
where $m$ denotes the number of row vectors of $W$ and $I$ represents the identity matrix. Basically, the penalty function $p_{orth}$ intends to make all the row vectors of $W$ matrix to be orthogonal to each other while having unit norm when a large scalar coefficient $\beta$ is selected. For the ideal case that $WW^T = I$, it is immediate that all the row vectors of $W$ are fully orthonormal.  Furthermore, all the eigenvalues of $WW^T$ are 1, leading to flat eigenvalue distributions. 

As mentioned in the introduction, employment of orthonormal regularisation needs manual-tuning of the the scalar coefficient $\beta$ in (\ref{equ:optimization}) to properly balance the importance of the objective function and the penalty term. Large $\beta$ value would slow down the learning process while small value would have little effect on the behaviours of the weight matrices.  Thus, it is highly beneficial to turn orthonormal regularisation into orthonormal normalisation to avoid or alleviate manual parameter-tuning. 
  



\subsection{General Framework of AON}
\label{subsec:AON}

Without loss of generality, we consider the input-output relationship under a weight matrix $W\in \mathbb{R}^{m\times n}$ at a particular neural layer. We drop the layer index for simplicity.  We would like to construct a function $h(W)\in \mathbb{R}^{m\times n}$ such that its row vectors are approximately orthonormal up to a positive scalar, which can be mathematically represented as 
\begin{align}
h(W)h(W)^T \approx I/s,
\label{equ:AON}
\end{align}
where $s>0$ represents the scalar, which remains to be specified. 
The output $z$ under $h(W)$ can be expressed as
\begin{align}
z =h(W)v, 
\label{equ:mat_multi}
\end{align}
where $v\in \mathbb{R}^n$ represents the output from one layer below right after a nonlinear activation function. 

Similarly to BN(or IterNorm) and WN(or CWN), to increase the representational power of the resulting DNN model, we introduce an additional scaling parameter for each element $z_i$ of the $z$ vector in (\ref{equ:mat_multi}), expressed as
\begin{align}
\tilde{z}_i =  \gamma_i z_i\quad i =1,\ldots, m. \label{equ:layer_scaling}
\end{align}
The resulting vector $\tilde{z}$ is then passed through a nonlinear function to obtain the output vector $v$ for the layer above. 

\begin{algorithm}[t]
\label{label:AON}
\caption{AON with Talyor expansion}
  \textbf{Input: } $\left\{\begin{array}{l}W\hspace{-0.3mm}\in \hspace{-0.3mm} \mathbb{R}^{m\times n}: \textrm{weight matrix. } \\
   u \hspace{-0.3mm}\in\hspace{-0.3mm} \mathbb{R}^m: \textrm{an eigenvector estimator. }  \end{array} \right.
  $. \\ 
 \textbf{Hyperparameters:} order $q$ of Talyor expansion. \\
 \textbf{Output:}  $h(W)$ and updated  eigenvector estimator $u$. \\
 calculate Talyor expansion $P_{q}(W)$  of $(WW^T)^{-\frac{1}{2}}$ up to order $q$. \\
  update $u$ and calculate the approximated spectral norm $\sigma_{P_q}$ of $P_{q}(W)W$ by using SN from \cite{Miyato18SN}. \\
  compute $h(W) = P_{q}(W)W/ \sigma_{P_q}$.
\end{algorithm}

Intuitively speaking, suppose there exists a DNN model that fits the training data sufficiently accurate under the mapping (\ref{equ:mat_multi})-(\ref{equ:layer_scaling}) and also approximately satisfies (\ref{equ:AON}) for all weight matrices. (\ref{equ:AON})-(\ref{equ:mat_multi}) together imply that at each neural layer, information of the input feature vector $v$ is encoded by all the orthonormal row vectors of $h(W)$ in a manner of equal importance.  By doing so,  it would avoid co-adaptation among the neurons that correspond to the elements of the output vector $z$. In this aspect, our new technique AON has a similar effect as the dropout \cite{Srivastava14Dropout} technique which is also designed to avoid co-adaptation for improving DNN generalisation capacity.  While dropout randomly drops neurons of the DNN model in the training process to avoid co-adaptation, AON imposes approximate orthonormality to the weight matrices. In brief, the two techniques follow different methodologies to reach the same goal. 

The scaling operation (\ref{equ:layer_scaling}) is able to suppress or enhance the information of $v$ associated with different row vectors of $h(W)$ via $z$ accordingly. That is, in addition to the nonlinear function, the scaling operation provides more freedom to control the information flow when training the DNN model. By doing so,  the representational power of the DNN model is naturally increased. 

Next we present the procedure for computing $h(W)$ in two steps. Firstly, we  attempt to obtain an approximation of $(WW^T)^{-\frac{1}{2}}$ assuming the matrix product $WW^T$ is nonsingular, denoted as
\begin{align}
P(W) \approx (WW^T)^{-\frac{1}{2}}. \label{equ:Taylor1}
\end{align}
We will consider the singular case later on. If a good approximation $P(W)$ exists, it is immediate that 
\begin{align}
&P(W)WW^TP(W)^T  \nonumber \\
&\approx (WW^T)^{-\frac{1}{2}}WW^T (WW^T)^{-\frac{1}{2}} \nonumber \\
&\approx I,
 \label{equ:P(W)_good}
\end{align} 
which implies that the row vectors of the matrix product $P(W)W$ are orthogonal to each other. As will be discussed in next subsection, we will make use of Taylor expansion to compute $P(W)$. 

Suppose $P(W)$ is obtained by Taylor approximation, our 2nd step is to scale the matrix product $P(W)W$ by using the SN technique developed in \cite{Miyato18SN}. We denote the spectral norm of $P(W)W$ as $\sigma_{P}$ after applying SN in \cite{Miyato18SN}. $h(W)$ can then be expressed as 
\begin{align}
h(W) = P(W)W/\sigma_{P}.
 \label{equ:h(W)}
\end{align}
Our motivation for scaling $P(W)W$ is that $P(W)$ may not always be a good approximation of  $(WW^T)^{-\frac{1}{2}}$, especially when $WW^T$ is singular. The SN technique ensures that the spectral norm of $h(W)$ in (\ref{equ:h(W)}) roughly equals to 1, which implicitly prevents $h(W)$ from growing of out control in the DNN training procedure. 

We note that the parameter $\sigma_{P}$ is a function of $P(W)W$, which is updated per training iteration to adapt to the changes of $W$.  As demonstrated in \cite{Miyato18SN}, $\sigma_{P}$ is estimated by using the power iteration method \cite{Golub00powerIter} instead of using SVD or other expensive operations. To do so, an estimator $u\in \mathbb{R}^{m}$ of the eigenvector corresponding to the maximum eigenvalue of $P(W)W$ is maintained during the whole training procedure.  When $W$ is updated by SGD based optimisation method, the vector $u$ is then updated accordingly using the power iteration method. The approximated spectral norm $\sigma_{P}$ is finally computed with the updated vector $u$ (see \textbf{Algorithm 1} for a brief summary).  

When conducting back-propagation per iteration, gradient of $\sigma_{P}$ w.r.t. $W$ is computed to preserve training stability. See \cite{Miyato18SN} for detailed gradient derivation for $\sigma_{P}$. Similarly, gradient of $P(W)$ w.r.t. $W$ is also computed when updating $W$ for the same purpose. 

Finally we reconsider (\ref{equ:AON}). Plugging (\ref{equ:h(W)}) into (\ref{equ:AON}), it is clear that the scalar $s$ approximately equals to $1/\sigma_{P}^2$ when (\ref{equ:P(W)_good}) holds with high accuracy. That is, the scalar $s$ is determined by the approximation $P(W)$ and the scaling operation for $P(W)W$. If Frobenius norm is used instead of SN when scaling $P(W)W$, the scalar $s$ is changed accordingly.

\subsection{Approximating $(WW^T)^{-\frac{1}{2}}$ by Taylor Expansion}

Given the square matrix $(WW^T)^{-\frac{1}{2}}$, we compute its Taylor expansion around the identity matrix. To do so, it is noted that the Taylor expansion of the corresponding scalar function $y(x)=x^{-\frac{1}{2}}$ around 1 can be expressed as 
\begin{align}
x^{-\frac{1}{2}} &\approx y(1) + \frac{y'(1)(x-1)}{1!} + \frac{y''(1)(x-1)^2}{2!} +\ldots \nonumber \\
	          & = 1 -\frac{1}{2}(x-1) +\frac{3}{8}(x-1)^2 +\ldots, \label{equ:scalar_Taylor}
\end{align}
where the expansion holds when $2>x>0$. As $x$ approaches to 1, the approximation becomes increasingly accurate for a particular gradient order. 
With (\ref{equ:scalar_Taylor}), the Taylor expansion of $(WW^T)^{-\frac{1}{2}}$ can be easily obtained by substituting $x$ in (\ref{equ:scalar_Taylor}) with $WW^T$, 
 which is given by \cite{BhatiaBook97Matrix}
\begin{align}
&P(W) \nonumber \\
& \approx I - \frac{1}{2}(WW^T-I) + \frac{3}{8}(WW^T-I)^2+\ldots. \label{equ:matrix_Taylor}
\end{align}
We use $P_q(W)$ to denote the Taylor approximation of $(WW^T)^{-\frac{1}{2}}$ up to order $q$. For example, $P_1(W) = \frac{3}{2}I- \frac{1}{2}WW^T$, when $T=1$. See \textbf{Algorithm} \ref{label:AON} for exploiting the notation $P_q(W)$ in computing $h(W)$.

In principle, the expansion (\ref{equ:matrix_Taylor}) holds under the condition 
\begin{align} 
0< | \lambda_i(WW^T) | <2, \label{equ:Taylor_cond}
\end{align}
 for all eigenvalues $\{\lambda_i\}$ of $WW^T$. That is, the matrix $WW^T$ has to be symmetric positive definite with its eigenvalues bounded within $(0, 2)$. In practice, it may happen that $WW^T$ is singular when the number of rows in $W$ is greater than the number of its columns. In this situation, our objective becomes to make the row vectors of $h(W)$ to be less correlated to each other as a relaxation of strict orthogonality. It is found empirically that the approximation (\ref{equ:matrix_Taylor}) together with the SN (\ref{equ:h(W)}) leads to high quality DNN models without the need to pay much attention to the boundness of the eigenvalues of $WW^T$. The SN  (\ref{equ:h(W)}) is crucial to ensure that the norm of $h(W)$ does not change dramatically. 

Next we briefly explain why we do not exploit Newton's iteration to approximate $(WW^T)^{-\frac{1}{2}}$. As is demonstrated in \cite{Huang19IterNorm} for the development of IterNorm, the degree of the polynomial of $x$ for the function $y=x^{-\frac{1}{2}}$ increases suplinearly when the iteration increases linearly. On the contrary, with regard to Taylor expansion (\ref{equ:scalar_Taylor}), the degree of the polynomial of $x$ increases linearly, which is more suitable for our application. 

\begin{remark}
To satisfy (\ref{equ:Taylor_cond}), we have also tried to first perform SN for $W$, and then compute $P(W/\sigma(W))W/\sigma(W)$ where $\sigma(W)$ denotes the spectral norm of $W$. It is found that by doing so, it accelerates the training speed at the cost of degraded validation performance.  Thus, it is preferable to first obtain the matrix product $P(W)W$ and then apply the SN technique.  
\end{remark}

\subsection{Training and Inference}
Similarly to existing normalisation methods, our AON can be easily incorporated into a DNN as an additional module. When employing AON for training and testing a DNN model in practice, two things are worthy being noticed. Firstly, the eigenvector estimator $u$ for each weight matrix $W$ (see \textbf{Algorithm 1}) in the considered DNN model needs to be maintained during the whole training procedure. The set of $u$ vectors in the model are updated at each training iteration while being fixed at the test stage. Secondly, calculation of $h(W)$ introduces computational overhead only at the training procedure. When the model is well-trained, $h(W)$ can then be computed once and stored as the model parameters without recomputing $h(W)$. the set of $u$ vectors can also be dropped to save storage space.     

We notice that the computational complexity of $h(W)$ and its back-propagation depends on the order $q$ for $P_q(W)$ in \textbf{Algorithm 1}. High order $q$ would naturally incur expensive computation. As will be discussed in the experimental part later on, increasing order $q$ does not consistently improve validation performance. There exists an proper $q$ value that keeps the correlation among the row vectors of $h(W)$ to a certain degree, which leads to high quality DNN models after the training procedure. In other words, it might not be necessary to totally remove the correlation among the row vectors of $h(W)$ even if it is doable. This is consistent with the observations made for IterNorm in \cite{Huang19IterNorm}. The authors of IterNorm found empirically that it is preferable to perform approximate-whitening of the covariance matrix of the internal features in a neural layer over full-whitening.  

\vspace{2mm}
\noindent\textbf{Convolutional Layer:}  Formulation (\ref{equ:optimization}) is for a densely connected DNN. Consider a weight tensor $W_c\in \mathbb{R}^{d_o\times d_i\times h\times w}$ from a CNN layer, where $d_i$ and $d_o$ indicate the numbers of input and output channels, and $h$ and $w$ represent the height and width of the CNN kernel. Following the procedure of CWN and SN \cite{Huang17CWN, Miyato18SN}, we reshape $W_c$ into a matrix $\hat{W}_c$ of size $d_o\times (d_ihw)$. The operation of AON is then performed over the reshaped matrix $\hat{W}_c$ to obtain $h(\hat{W}_c)$, which is then reshaped back to be of size $d_o\times d_i\times h\times w$.

\section{Experiments}

\subsection{Experimental setup}

In the experiments, we consider training the VGG16 network over the CIFAR10 and CIFAR100 datasets using different normalisation or regularisation techniques. As the name suggests, CIFAR10 and CIFAR100 correspond to two classification problems of 10 and 100 classes, respectively. The implementation of the training and validation procedures were conducted based on the \footnote{https://github.com/huangleiBuaa/IterNorm.}{open source} for IterNorm (see \cite{Huang19IterNorm}), which was implemented over the pytorch platform. In brief, SGD with momentum was employed for training each network configuration, where the momentum was set to be 0.9. The maximum number of epochs was 160. The initial learning rate was 0.1, and scheduled to be divided by 2 at 60 and 120 epochs sequentially.  To alleviate the effect of the randomness in the training process, five experimental repetitions were conducted for each network configuration. 

The experiments were conducted in two steps. Firstly, we perform ablation studies on AON to find out the effect of order $q$ in $P_q(W)$ on its performance. We then conduct performance comparison of five network configuration based on different combinations of normalisation or regularisation techniques from literature. Our primary interest is the validation performance gain due to the introduction of AON compared to orthonormal regularisation in (\ref{equ:orthnorm}).

\subsection{Ablation studies on performance of AON}

In the first experiment, we study how the order $q$ of $P_q(W)$ (see \textbf{Algorithm 1}) affects the performance of AON by training VGG16 over CIFAR10.  The BN technique is utilised together with AON by default to accelerate the training speed. Three $q$ values were considered, which are $q=0$, 2, and $4$. For the special case that $q=0$, $P_q(W)=I$. In this situation, the function $h(W)$ in (\ref{equ:h(W)}) is degenerated to $h(W)=W/\sigma(W)$, where $\sigma(W)$ represents the spectral norm of $W$ obtained by applying SN in \cite{Miyato18SN}. That is, when $q=0$, AON reduces to SN.
In other words, SN is a special case of AON when the Taylor approximation is removed in AON. 

Table~\ref{tab:AON_val_acc_time} displays the validation performance and the average training time for the three different $q$ values of AON. It is observed that as $q$ increases, the training time (per epoch) also increases as expected. When it comes to validation performance, it is clear that the setup $q=2$ outperforms $q=0$ and $q=4$ by a noticeable gain. Furthermore, the performance of $q=4$ is also better than that of $q=0$, indicating that approximate orthonormalisation indeed improves network generalisation.
\begin{table}[h]
\caption{\small  Effect of order $q$ on the performance of AON for training VGG16 over CIFAR10. Each validation accuracy is obtained from  five experimental repetitions per $q$ value. The training time is obtained by averaging the running time from the first 30 epochs.
 } 
\label{tab:AON_val_acc_time}
\centering
\begin{tabular}{|c|c|c|c|}
\hline
&\hspace{-1mm} {{\scriptsize BN+AON(q=0)}}\hspace{-1mm}
& \hspace{1mm}{\scriptsize  BN+AON(q=2)}  \hspace{-1mm}
& \hspace{-1mm}{\scriptsize  $\begin{array}{c}\textrm{BN+AON(q=4)} \end{array}$}  \hspace{-1mm} 
 \\
\hline  
\hspace{-2mm}{\footnotesize  $\begin{array}{c}\textrm{validation} \\ \textrm{accuracy (\%)} \end{array} $} \hspace{-2mm} & \hspace{-1mm} \footnotesize{92.28$\pm 0.12$} \hspace{-1mm}&\hspace{-1mm} \footnotesize{\textbf{93.51}$\pm 0.05$} \hspace{-1mm}&\hspace{-1mm} \footnotesize{93.21$\pm 0.10$} \\ 
\hline
\hspace{-2mm}{\footnotesize  $\begin{array}{c}\textrm{training time} \\ \textrm{per epoch (s)} \end{array} $} \hspace{-2mm} & \hspace{-1mm} \footnotesize{20.0} \hspace{-1mm}&\hspace{-1mm} \footnotesize{27.1} \hspace{-1mm}&\hspace{-1mm} \footnotesize{32.5} \\ 
\hline
\end{tabular}
\vspace{0mm}
\end{table}

\begin{figure}[h]
\centering
\includegraphics[width=80mm]{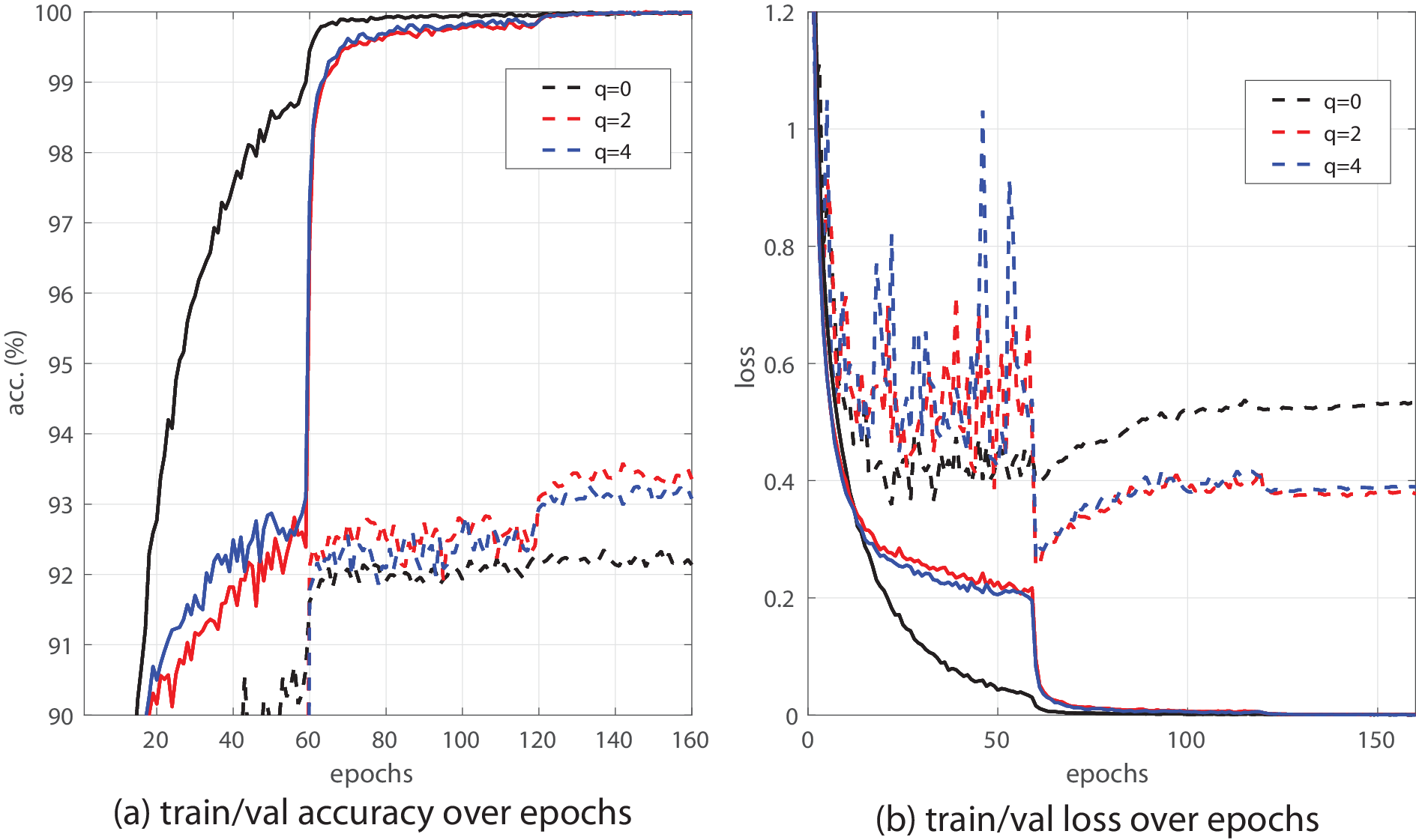}
\caption{ Effect of order $q$ on the performance of AON for training VGG16 over CIFAR10. The solid curves are for training performance while the dashed ones are for validation performance. The subplots (a) and (b) display the effects of different order $q$ on the training and validation performance, respective.  The curve for each configuration is selected from five experimental repetitions, which gives the highest validation accuracy. } 
\label{fig:VGG_orderq}
\end{figure}

Fig.~\ref{fig:VGG_orderq} visualises the trajectories of training/validation performance of AON over the 160 epochs. It is seen that the performance of $q=2$ and $q=4$ is comparable from an overall perspective. Given the fact (see Table \ref{tab:AON_val_acc_time}) that the computational complexity for $q=2$ is considerably lower than that for $q=4$, $q=2$ is a preferable setup. Based on the above observations, we will use $q=2$ for AON in the following experiments. 

For completeness, we provide the explicit expression for $P_2(W)$ when $q=2$, given by
\begin{align}
P_2(W) = 1.875 I - 1.25 WW^T + 0.375 (WW^T)^2. \label{equ:P_2}
\end{align}
We note that $P_2(W)$ is a 2nd order polynomial of $WW^T$, which is similar to the expression $P_{orth}(W)$ in (\ref{equ:orthnorm}) for orthonormal regularisation. One main difference is that the polynomial coefficients in $P_2(W)$ are derived based on the Taylor expansion of $(WW^T)^{-\frac{1}{2}}$ around the identity matrix. Interested readers can also work out the expressions for high order of $P_q(W)$ based on (\ref{equ:scalar_Taylor})-(\ref{equ:matrix_Taylor}) for advanced study.

\begin{figure*}[t!]
\centering
\includegraphics[width=160mm]{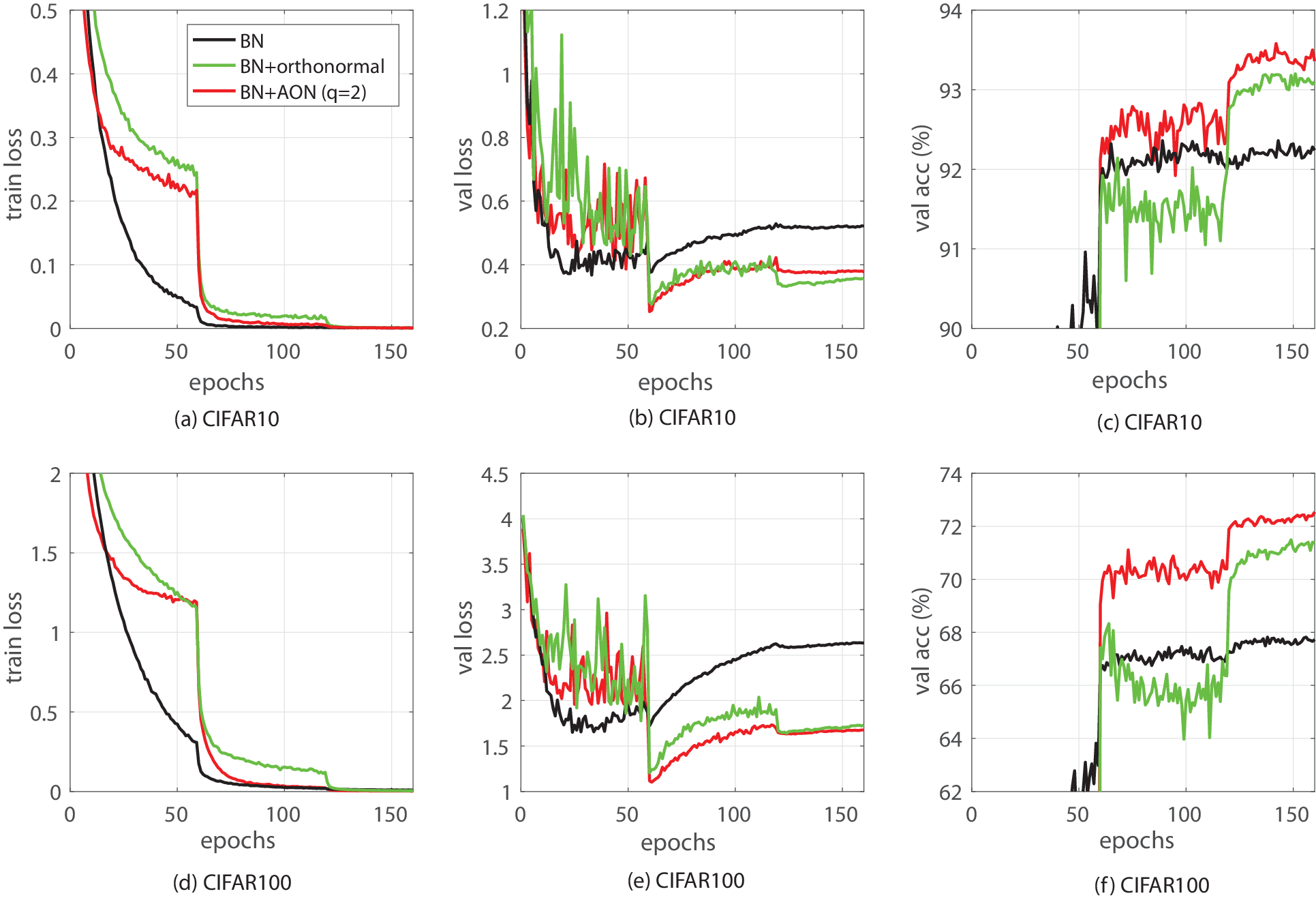}
\caption{ Performance visualisation of BN, BN+orthonormal, and BN+AON(q=2) for training VGG16 over both CIFAR10 and CIFAR100, where  \emph{orthonormal} refers to orthonormal regularisation. The curve for each configuration is selected from five experimental repetitions, which gives the highest validation accuracy. } 
\label{fig:VGG}
\end{figure*}

\subsection{Performance comparison}

In this experiment, we tested five configurations for VGG16, which are BN,  BN+CWN, IterNorm, BN+orthonormal, and  $\textrm{BN+AON(q=2)}$, where \emph{orthonormal} refers to orthonormal regularisation defined by (\ref{equ:orthnorm}). The scalar coefficient $\beta$ in (\ref{equ:optimization}) for orthonormal regularisation was set to be 10.0, which leads to better performance than $\beta=1$ and $\beta=20$.  In addition, the hyper-parameters for IterNorm were set according to the suggestions provided in \cite{Huang19IterNorm}.

\vspace{1.5mm}
\noindent\textbf{For CIFAR10:} Table \ref{tab:val_acc_cifar10} shows the validation accuracy and (averaging) training time of the five configurations for training VGG16 over CIFAR10. It is clear that the configuration BN+AON(q=2) yields the best validation performance while BN alone gives the worst performance.  BN+CWN and IterNorm perform better than BN alone but their performance is inferior to that of BN+orthonormal and BN+AON (q=2). Furthermore, BN+AON(q=2) has a noticeable gain compared to BN+orthonormal.  

\begin{table}[h]
\caption{\small Performance comparison of five configurations for training the VGG16 on CIFAR10.  Each Validation accuracy is obtained from five experimental repetitions per configuration. The training time is obtained by averaging the running time from the first 30 epochs.
 } 
\label{tab:val_acc_cifar10}
\centering
\begin{tabular}{|c|c|c|c|}
\hline
 &\hspace{0mm} {{\scriptsize BN}}\hspace{0mm}
& \hspace{0mm}{\scriptsize  BN+CWN}  \hspace{0mm}
& \hspace{0mm}{\scriptsize  $\begin{array}{c}\textrm{IterNorm} \end{array}$}  \hspace{0mm} 
 \\
\hline  
\hspace{-2mm}{\footnotesize  $\begin{array}{c}\textrm{validation} \\ \textrm{accuracy (\%)} \end{array} $} \hspace{-2mm}& \footnotesize{92.22$\pm 0.09$} & \footnotesize{92.68$\pm 0.08$} & \footnotesize{92.66$\pm 0.21$} \\ 
\hline
\hspace{-2mm}{\footnotesize  $\begin{array}{c}\textrm{training time} \\ \textrm{per epoch (s)} \end{array} $} \hspace{-2mm} & \hspace{-1mm} \footnotesize{17.2} \hspace{-1mm}&\hspace{-1mm} \footnotesize{20.7} \hspace{-1mm}&\hspace{-1mm} \footnotesize{37.1} \\ 
\hline
\hline
& \hspace{0mm} {\scriptsize $\textrm{BN+orthornormal}$} &   {\scriptsize $\textrm{BN+AON(q=2)}$}  &   \\
 \cline{1-3}
 \hspace{-2mm}{\footnotesize  $\begin{array}{c}\textrm{validation} \\ \textrm{accuracy (\%)} \end{array} $} \hspace{-2mm} & \footnotesize{93.09$\pm0.09$}  & \footnotesize{$\textbf{93.51}\pm 0.05$}  & \\
 \cline{1-3}
 \hspace{-2mm}{\footnotesize  $\begin{array}{c}\textrm{training time} \\ \textrm{per epoch (s)} \end{array} $} \hspace{-2mm} & \hspace{-1mm} \footnotesize{21.1} \hspace{-1mm}&\hspace{-1mm} \footnotesize{27.1} \hspace{-1mm}& \\ 
\hline
\end{tabular}
\vspace{0mm}
\end{table}

Considering the training time in Table  \ref{tab:val_acc_cifar10}, IterNorm is most expensive in running time due to the internal Newton's Iteration while BN alone costs least running time. BN+AON(q=2) is slightly slower than BN+orthonormal because of the computation of $h(W)$ and its back-propagation. From a high-level perspective, introduction of the above normalization or regularisation techniques into the DNN model increases the training time while at the same time, improves the generalisation capacity of the neural network. 

\vspace{1.5mm}
\noindent\textbf{For CIFAR100:} Based on the experimental observations from CIFAR10 above, we tested three configurations when training VGG16 over CIFAR100, which are BN, BN+orthonormal, and BN+AON(q=2).
  
\begin{table}[h]
\caption{\small  Performance comparison of three configurations for training the VGG16 on CIFAR100. Each Validation accuracy is obtained from five experimental repetitions per configuration. The training time is obtained by averaging the running time from the first 30 epochs.} 
\label{tab:val_acc_cifar100}
\centering
\begin{tabular}{|c|c|c|c|}
\hline
&\hspace{-1mm} {{\scriptsize BN}}\hspace{-1mm}
& \hspace{1mm}{\scriptsize  BN+orthornormal}  \hspace{-1mm}
& \hspace{-1mm}{\scriptsize  $\begin{array}{c}\textrm{BN+AON(q=2)} \end{array}$}  \hspace{-1mm} 
 \\
 \hline  
\hspace{-2mm}{\footnotesize  $\begin{array}{c}\textrm{validation} \\ \textrm{accuracy (\%)} \end{array} $} \hspace{-2mm} & \hspace{-1mm} \footnotesize{67.20$\pm 0.33$} \hspace{-1mm}&\hspace{-1mm} \footnotesize{70.92$\pm 0.34$} \hspace{-1mm}&\hspace{-1mm} \footnotesize{\textbf{72.10}$\pm 0.44$} \\ 
\hline
\hspace{-2mm}{\footnotesize  $\begin{array}{c}\textrm{training time} \\ \textrm{per epoch (s)} \end{array} $} \hspace{-2mm} &\hspace{-1mm} \footnotesize{17.3} \hspace{-1mm}&\hspace{-1mm} \footnotesize{21.1} \hspace{-1mm}&\hspace{-1mm} \footnotesize{27.2}  \\
\hline
\end{tabular}
\vspace{0mm}
\end{table}
  
The performance of the three configurations is summarised in Table \ref{tab:val_acc_cifar100}. One observes that the validation performance of BN+orthonormal and BN+AON (q=2) is significantly better than that of BN alone. Similarly to that of CIFAR10, BN+AON (q=2) performs better than BN+orthonormal. In addition, the  (averaging) training times of the three configurations are consistent across the two datasets as shown in Table~\ref{tab:val_acc_cifar10} and \ref{tab:val_acc_cifar100}.

\vspace{1.5mm}
\noindent\textbf{Performance visualisation for CIFAR10 and CIFAR100:} Fig.~\ref{fig:VGG} displays the convergence results of thee three configurations BN, BN+orthonormal, and BN+AON(q=2). The results for BN+CWN and IterNorm are omitted to avoid distraction.  Each curve in the plot is selected from five experimental repetitions which gives the highest validation accuracy.  

It is clearly seen from the subplots (a) and (d) that BN+AON(q=2) converges slower than $BN$ but faster than BN+orthonormal w.r.t. number of epochs. Considering the validation loss and accuracy, both BN+AON(q=2) and BN+orthonormal outperform BN considerably. Furthermore, the validation accuracy of BN+AON(q=2) is noticeably better than that of BN+orthonormal for the two datasets.   

To briefly summarise, BN+AON performs consistently better than other tested normalisation or regularisation techniques for both CIFAR10 and CIFAR100. This suggests that AON manages to avoid co-adaption in the neural layers of a DNN model, thus successfully improving its generalisation capacity.  While BN+orthonormal needs manual parameter-tuning to balance the objective function and the penalty term, AON incorporates approximate orthonormality into the DNN model in a seamless manner. 

\section{Conclusions and Future works}
In this paper, we have proposed a new weight normalisation technique, named as \emph{approxiamted orthonormal normalisation} (AON).  AON is designed to turn orthonormal regularisation into orthonormal normalisation to make the training procedure simpler and cleaner. Given a weight matrix $W$ of a neural layer, the basic idea of AON is to construct a function $h(W)$ such that its row vectors are approximately orthogonal to each other. To do so, the key step is to obtain an approximation of $(WW^T)^{-\frac{1}{2}}$ using Taylor expansion, which is then multiplied by $W$. The spectral normalisation (SN) technique is used to normalise the obtained matrix product to obtain $h(W)$. Compared to the orthonormal regularisation technique, AON avoids introducing additional penalty term to the objective function, making the training process neat. Experimental results indicate that AON outperform not only orthonormal regularisation but also other normalisation techniques such as CWN and IterNorm. 

One future research direction for AON is to study the possibility of constructing a more effective function $h(W)$ by exploiting, for instance, different forms of Taylor expansion and/or different scaling operations in addition to SN. 

As AON and IterNorm operate on weight matrices and internal features of input data, respectively, one can consider combining AON and IterNorm in a proper way to improve the system performance to a higher level. This could be realized by conducting decorrelation in both weight matrices and internal features of a neural layer simultaneously.

%
\IEEEpeerreviewmaketitle

\ifCLASSOPTIONcaptionsoff
  \newpage
\fi


\begin{thebibliography}{10}
\providecommand{\url}[1]{#1}
\csname url@samestyle\endcsname
\providecommand{\newblock}{\relax}
\providecommand{\bibinfo}[2]{#2}
\providecommand{\BIBentrySTDinterwordspacing}{\spaceskip=0pt\relax}
\providecommand{\BIBentryALTinterwordstretchfactor}{4}
\providecommand{\BIBentryALTinterwordspacing}{\spaceskip=\fontdimen2\font plus
\BIBentryALTinterwordstretchfactor\fontdimen3\font minus
  \fontdimen4\font\relax}
\providecommand{\BIBforeignlanguage}[2]{{%
\expandafter\ifx\csname l@#1\endcsname\relax
\typeout{** WARNING: IEEEtran.bst: No hyphenation pattern has been}%
\typeout{** loaded for the language `#1'. Using the pattern for}%
\typeout{** the default language instead.}%
\else
\language=\csname l@#1\endcsname
\fi
#2}}
\providecommand{\BIBdecl}{\relax}
\BIBdecl

\bibitem{Nair10ReLU}
V.~Nair and G.~E. Hinton, ``{Rectified Linear Units Improve Restricted
  Boltzmann Machines},'' in \emph{Proceedings of the 27th International
  Conference on Machine Learning,}, 2010.

\bibitem{He15WeiInti}
K.~He, X.~Zhang, S.~Ren, and J.~Sun, ``{Delving Deep into Rectifiers:
  Surpassing Human-Level Performance on Imagenet Classification},'' in
  \emph{Proceedings of the IEEE international conference on computer vision},
  2015, pp. 1026--1034.

\bibitem{Sutskever13NAG}
H.~Sutskever, J.~Martens, G.~Dahl, and G.~Hinton, ``{On the importance of
  initialization and momentum in deep learning},'' in \emph{International
  conference on Machine Learning (ICML)}, 2013.

\bibitem{He15ResNet}
K.~He, X.~Zhang, S.~Ren, and J.~Sun, ``{Deep Residual Learning for Image
  Recognition},'' in \emph{IEEE conference on Computer Vision and Pattern
  Recognition (CVPR)}, 2015.

\bibitem{Zagoruyko16WideResNet}
S.~Zagoruyko and N.~Komodakis, ``{Wide Residual Networks},''
  arXiv:1605.07146v4, 2016.

\bibitem{Huang17DenseNet}
G.~Huang, Z.~Liu, L.~ver~der Maaten, and K.~Q. Weinberger, ``{Densely Connected
  Convolutional Networks},'' in \emph{IEEE conference on Computer Vision and
  Pattern Recognition (CVPR)}, 2017.

\bibitem{Ronneberger15Unet}
O.~Ronneberger, P.~Fischer, and T.~Brox, ``{U-Net: Convolutional Networks for
  Biomedical Image Segmentation},'' arXiv:1505.04597 [cs.CV], 2015.

\bibitem{Kingma17}
D.~P. Kingma and J.~L. Ba, ``{Adam: A Method for Stochastic Optimization},''
  arXiv preprint arXiv:1412.6980v9, 2017.

\bibitem{Reddi18Amsgrad}
S.~K. S.~J.~Reddi and S.~Kumar, ``{On the Convergence of Adam and Beyond},'' in
  \emph{International conference on Learning Representations (ICLR)}, 2018.

\bibitem{Duchi11AdaGrad}
J.~Duchi, E.~Hazan, and Y.~Singer, ``{Adaptive Subgradient Methods for Online
  Learning and Stochastic Optimization},'' \emph{Journal of Machine Learning
  Research}, vol.~12, pp. 2121--2159, 2011.

\bibitem{Tieleman12RMSProp}
T.~Tieleman and G.~Hinton, ``{Lecture 6.5-RMSProp: Divide The Gradient by a
  Running Average of Its Recent Magnitude},'' COURSERA: Neural networks for
  machine learning, pp. 26--31, 2012.

\bibitem{Ioffe15BN}
S.~Ioffe and C.~Szegedy, ``{Batch normalization: Accelerating Deep Network
  Training by Reducing Internal Covariate Shift},'' \emph{volume 37 of JMLR
  Proceedings}, pp. 448--456, 2015.

\bibitem{Ioffe17BRN}
S.~Ioffe, ``{Batch Renormalization: Towards Reducing Minibatch Dependence in
  Batch-Normalized Models},'' in \emph{Advances in Neural Information
  Processing}, 2017.

\bibitem{Ba16LN}
G.~E.~H. J.~L.~Ba, J. R.~Kiros, ``{Layer Normalization},'' arXiv:1607.06450
  [stat.ML], 2016.

\bibitem{Ulyanov17InstanceNorm}
D.~Ulyanov, A.~Vedaldi, and V.~Lempitsky, ``{Instance Normalization: The
  Missing Ingredient for Fast Stylization},'' arXiv:1607.08022 [cs.CV], 2017.

\bibitem{Wu18GroupNorm}
Y.~Wu and K.~He, ``{Group normalization},'' in \emph{European Conference on
  Computer Vision (ECCV)}, 2018.

\bibitem{Huang18DBN}
L.~Huang, D.~Yang, B.~Lang, and J.~Deng, ``{Decorrelated Batch
  Normalization},'' in \emph{Conference on Computer Vision and Pattern
  Recognition(CVPR)}, 2018, pp. 791--800.

\bibitem{Huang19IterNorm}
L.~Huang, Y.~Zhou, F.~Zhu, L.~Liu, and L.~Shao, ``{Iterative Normalization:
  Beyond Standardization towards Efficient Whitening},'' in \emph{Conference on
  Computer Vision and Pattern Recognition(CVPR)}, 2019, pp. 4874--4883.

\bibitem{Klambauer17SNN}
G.~Klambauer, T.~Unterthiner, A.~Mayr, and S.~Hochreiter, ``{Self-Normalizing
  Neural Networks},'' in \emph{31st Conference on Nueral Information Processing
  Systems (NIPS)}, 2017.

\bibitem{Salimans16WN}
D.~P.~K. T.~Salimans, ``{Weight Normalization: A Simple Reparameterization to
  Accelerate Training of Deep Neural Networks},'' arXiv:1602.07868 [cs.LG],
  2016.

\bibitem{Huang17CWN}
L.~Huang, X.~Liu, Y.~Liu, B.~Lang, and D.~Tao, ``{Centered Weight Normalization
  in Accelerating Training of Deep Neural Networks},'' in \emph{International
  Conference on Computer Vision}, 2017, pp. 2803--2811.

\bibitem{Miyato18SN}
T.~Miyato, T.~Kataoka, M.~Koyama, and Y.~Yoshida, ``{Spectral Normalization for
  Generative Adversarial Networks},'' in \emph{ICLR}, 2018.

\bibitem{Hoffer18N}
E.~Hoffer, R.~Banner, I.~Golan, and D.~Soudry, ``{Norm Matters: Efficient and
  Accurate Normalization Schemes in Deep Networks},'' arXiv:1803.01814
  [stat.ML], 2018.

\bibitem{Brock16orth}
A.~Brock, T.~Lim, J.~M. Ritchie, and N.~Westona, ``{Neural photo editing with
  introspective adversarial networks},'' arXiv preprint arXiv:1609.07093, 2016.

\bibitem{Jia19ODNN}
K.~Jia, S.~Li, Y.~Wen, T.~Liu, and D.~Tao, ``{Orthogonal Deep Neural
  Networks},'' arXiv:1905.05929v2 [cs.LG], 2019.

\bibitem{Bansal18OR}
N.~Bansal, X.~Chen, and Z.~Wang, ``{Can We Gain More from Orthogonality
  Regularizations in Training Deep CNNs?}'' in \emph{Advances in Neural
  Information Processing}, 2018.

\bibitem{Simonyan16DCNN}
K.~Simonyan and A.~Zisserman, ``{Very Deep Convolutional Networks for
  Large-Scale Image Recognition},'' in \emph{International conference on
  Learning Representations (ICLR)}, 2016.

\bibitem{Yoshida17SNR}
Y.~Yoshida and T.~Miyato, ``{Spectral Norm Regularization for Improving the
  Generalizability of deep learning},'' arXiv preprint arXiv:1705.10941, 2017.

\bibitem{Krogh92weightdecay}
A.~Krogh and J.~A. Hertz, ``{A Simple Weight Decay Can Improve
  Generalization},'' in \emph{Advances in Neural Information Processing}, 1992,
  pp. 950--957.

\bibitem{Srivastava14Dropout}
N.~Srivastava, G.~Hinton, A.~Krizhevsky, I.~Sutskever, and R.~Salakhutdinov,
  ``{Dropout: A Simple Way to Prevent Neural Networks from Overfitting},''
  \emph{Journal of Machine Learning Research}, pp. 1929--1958, 2014.

\bibitem{Golub00powerIter}
G.~H. Golub and H.~A.~V. der Vorst, ``{Eigenvalue computation in the 20th
  century},'' \emph{Journal of Computational and Applied Mathematics}, vol.
  123, no.~1, pp. 35--65, 2000.

\bibitem{BhatiaBook97Matrix}
R.~Bhatia, \emph{{Matrix Analysis}}.\hskip 1em plus 0.5em minus 0.4em\relax
  Springer, 1997.

\end{thebibliography}
\end{document}